\documentclass[runningheads]{llncs}
\usepackage{graphicx}
\usepackage{amsmath,amssymb} 
\usepackage{color}
\usepackage{textcomp}

\begin{document}
\pagestyle{headings}
\mainmatter

\def\ACCV20SubNumber{281}  

\title{Class-Wise Difficulty-Balanced Loss for Solving
	Class-Imbalance} 
\titlerunning{CDB Loss for Solving Class-Imbalance}
%
\author{Saptarshi Sinha\orcidID{0000-0002-5207-1551} \and
Hiroki Ohashi \and
Katsuyuki Nakamura\orcidID{0000-0002-8074-2279}}
\authorrunning{S. Sinha et al.}
%
\institute{Hitachi, Ltd. Research \& Development Group, Tokyo, 185-8601 Japan\\
\email{\{saptarshi.sinha.hx, hiroki.ohashi.uo, katsuyuki.nakamura.xv\}@hitachi.com}\\
}

\maketitle

\begin{abstract}
Class-imbalance is one of the major challenges in real world datasets, where a 
few classes (called majority classes) constitute much more data
samples than the rest (called minority classes). Learning deep neural networks
using such datasets leads to performances that are typically biased towards the
majority classes. Most of the prior works try to solve class-imbalance by 
assigning more weights to the minority classes in various manners (e.g., data
re-sampling, cost-sensitive learning). However, we argue that the number of
available training data may not be always a good clue to determine the weighting 
strategy because some of the minority classes might be sufficiently represented 
even by a small number of training data. Overweighting samples of such
classes can lead to drop in the model’s overall performance. We claim that the
‘difficulty’ of a class as perceived by the model is more important to determine
the weighting. In this light, we propose a novel loss function named Class-wise
Difficulty-Balanced loss, or CDB loss, which dynamically distributes weights
to each sample according to the difficulty of the class that the sample belongs
to. Note that the assigned weights dynamically change as the ‘difficulty’ for
the model may change with the learning progress. Extensive experiments are
conducted on both image (artificially induced class-imbalanced MNIST,
long-tailed CIFAR and ImageNet-LT) and video (EGTEA) datasets. The results show 
that CDB
loss consistently outperforms the recently proposed loss functions on 
class-imbalanced datasets irrespective of the data type (i.e., video or image).
\end{abstract}

\section{Introduction}
\label{intro}

Since the advent of Deep Neural Networks (DNNs), we have seen significant 
advancement in computer vision research. One of the reasons behind this success 
is
the wide availability of large-scale annotated image (e.g., MNIST \cite{MNIST_data}, CIFAR 
\cite{CIFAR}, ImageNet \cite{Imagenet}) and video (e.g., Kinetics \cite{Kinetics}, Something-Something \cite{Something-something}, UCF \cite{UCF}) datasets.
But unfortunately, most of the commonly used datasets do not resemble the 
real world data in a number of ways. As a result, performance of state-of-the-art
DNNs drop significantly in real-world use-cases. One of the major challenges in 
most
real-world datasets is the class-imbalanced data distribution with significantly 
long
tails, i.e., a few classes (also known as ‘majority classes’) have much higher 
number
of data samples compared to the other classes (also known as ‘minority 
classes’).
When DNNs are trained using such real-world datasets, their performance gets 
biased towards the majority classes, i.e., they perform highly for the majority 
classes
and poorly for the minority classes.

Several recent works have tried to solve the problem of 
class-imbalanced training data. Most of the prior solutions can be fairly classified 
under 3 categories :- (1) Data re-sampling techniques \cite{SMOTE,Diversity_based_oversampling,Undersampling_1} 
(2) Metric learning and knowledge transfer \cite{deepmetriclearning,metriclearning2,metriclearning3,metalearning1}
(3) Cost-sensitive 
learning methods \cite{FOCALloss,Class_balancedloss,Equalizationloss,Gradient_harmonised}. Data re-sampling techniques try to balance the number of data 
samples between the majority and minority classes by either over-sampling from 
the minority classes or under-sampling from the majority classes or using both. 
Generating synthetic data samples for minority classes \cite{SMOTE,BorderlineSMOTE,SafelevelSMOTE} from given 
data is another
re-sampling technique that tries to increase the number of minority class 
samples.
Since the performance of a DNN depends entirely on it’s ability to learn to 
extract
useful features from data, “what training data is seen by the DNN” is a very 
important
concern. In that context, data re-sampling strategies introduce the risks of 
losing important training data samples by under-sampling from majority classes 
and network
overfitting due to over-sampling minority classes. Metric-learning 
\cite{deepmetriclearning,metriclearning2}, on the other hand, aims to learn an 
appropriate representation function that embeds data to a feature space, where 
the mutual relationships among the data (e.g., similarity/dissimilarity) are 
preserved. It has the risk of learning a biased representation function that has 
learned more from the majority classes. Hence some works  \cite{metriclearning3} 
tend to use sampling techniques with metric learning, which still faces the 
problems of sampling, as discussed above. Few recent researches have also tried 
to transfer knowledge from the majority classes to the minority classes by 
adding an external memory \cite{OLTR}, which is non-trivial and expensive. Due 
to these 
concerns, the 
work
in this paper focuses on cost-sensitive learning approaches. Cost-sensitive 
learning
methods penalize the DNN higher for making errors on certain samples compared
to others. They achieve this by assigning weights to different samples using 
various
strategies. Typically, most prior cost-sensitive learning strategies \cite{Inverse_freq1,Smoothed_inv,Class_balancedloss} 
assume
that the minority classes are always weakly represented. They ensure that the 
samples of the minority class get higher weights so that the DNN can be 
penalized more
for making mistakes on the minority class samples. One such popular strategy is 
to
distribute weights in inverse proportion to the class frequencies \cite{Inverse_freq1,Inverse_freq2}. 
However, certain minority classes might be fairly represented by a small amount of 
samples.
\begin{figure}[t]
\centering
\includegraphics[width=\linewidth]{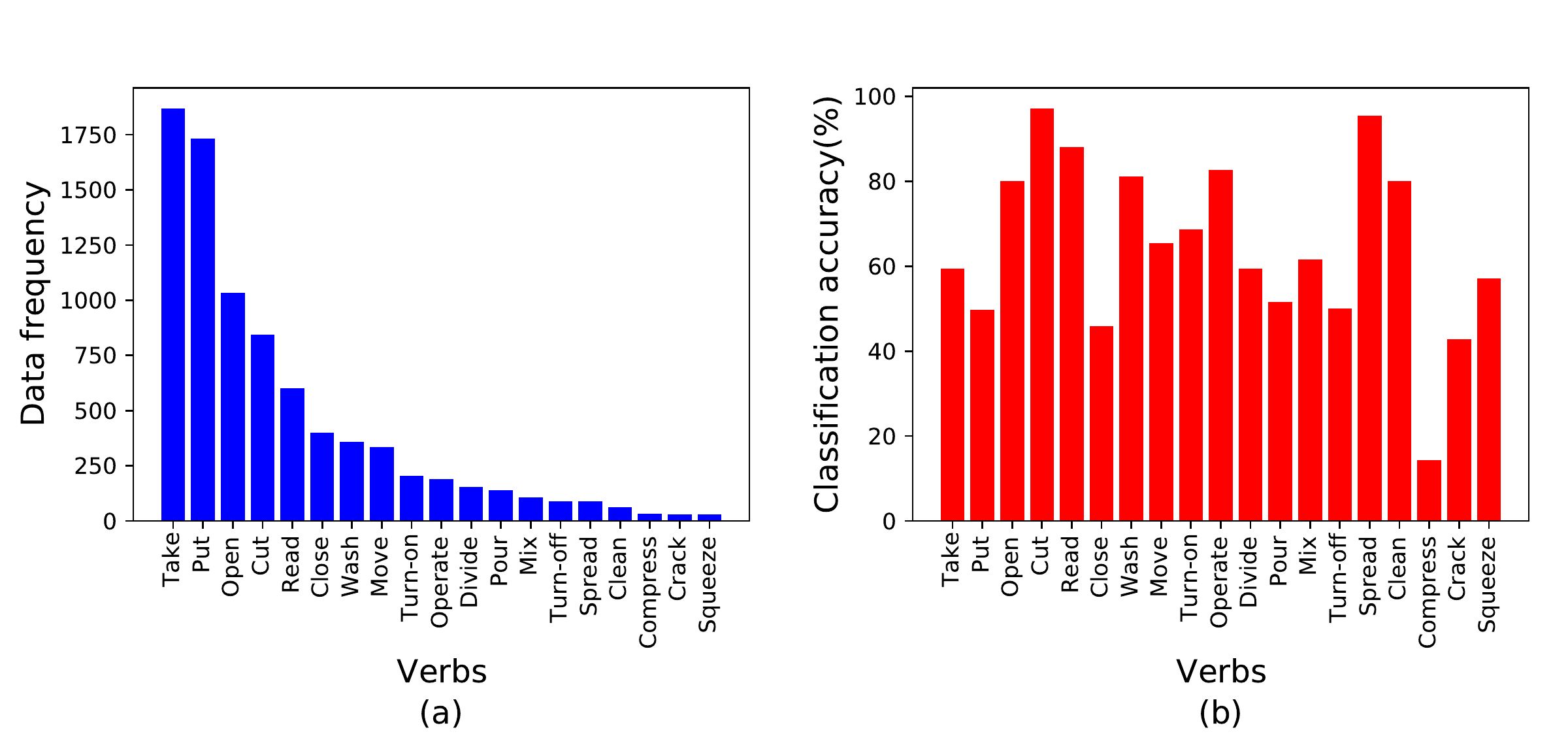}
\caption{
(a) Class-imbalanced data distribution of EGTEA dataset.
(b) Class-wise classification accuracies of a 3D-ResNeXt101
trained on the imbalanced EGTEA dataset using unweighted softmax cross-entropy loss function. It is interesting to notice that even though classes like ‘Clean’ and ‘Spread’ have very small
number of data samples, the classifier finds it relatively easier to learn such classes compared
to certain densely populated classes such as ‘Take’ and ‘Put’. Therefore it is not obvious to
assume that the sparsely populated classes will always be the most weakly represented.
}
\label{fig:egteafreq}
\end{figure}

Fig. \ref{fig:egteafreq} gives an example of a class-imbalanced dataset where certain classes such 
as
‘clean’ and ‘spread’ are sparsely populated but can easily be learned to 
generalize by
the classifier. Overweighting samples of such classes might lead to biasing the 
DNN’s
performance. In such situations, number of available training data per class 
might
not be a good clue to determine sample weights.

Instead, we claim that the ‘difficulty’ of a class as perceived by the DNN might 
be
a more important and helpful clue for weight assignment. The concept of 
‘difficulty’
has been previously used by some sample-level weight assigning techniques such
as focal loss \cite{FOCALloss} and GHM \cite{Gradient_harmonised}. They reweight each sample individually by 
increasing weights for hard samples and reducing weights for easy samples. The 
increasing
popularity of focal loss \cite{FOCALloss} in class-imbalanced classification tasks is based 
on the
assumption that minority classes should have more hard samples compared to 
majority classes. The assumption does stand true if we compare the proportion of 
hard
samples for the minority and majority classes. But, in terms of absolute number 
of
hard samples, the majority classes still might surpass the minority classes 
simply because they have much more data samples than the minority classes. In 
such cases,
giving high weights to all hard samples irrespective of their classes might 
overweight
the majority classes and therefore still bias the performance of the DNN. We 
believe
that the above drawback can be solved by considering class-level difficulty 
rather
than sample-level. To the best of our knowledge, ours is the first work to 
introduce
the concept of class-level difficulty for solving class-imbalance. Based on the 
analysis, we develop a novel weighting strategy that dynamically re-balances the 
loss for
each sample based on the instantaneous difficulty of it’s class as perceived by 
a DNN.

Such a strategy measures the instantaneous difficulty of each class without 
relying
on any prior assumptions and then dynamically assigns weights to the samples of
the class in proportion to the difficulty of the class. Extensive experiments on 
multiple datasets indicate that our class-difficulty based dynamic weighting 
strategy can
provide a significant improvement in the performance of the commonly used loss
functions under class-imbalanced situations.

The key contributions of this paper can be summarized as follows: (1) We propose 
a way to measure the dynamic difficulty of each class during training and use
the class-wise difficulty scores to re-balance the loss for each sample, thereby 
giving a
class-wise difficulty-balanced (CDB) loss. (2) We show that using our weighting 
strategy can give commonly used loss functions (e.g., cross-entropy) a 
significant boost
in performance on multiple class-imbalanced datasets. We conduct experiments on
both image and video datasets and find that our weighting strategy works well 
irrespective of the data type. Our research on quantifying the dynamic 
difficulty of the
classes and using it for weight distribution might prove useful for researchers 
focusing on class-imbalanced datasets.
\section{Related Works}
\label{sec:relatedworks}

As discussed in section \ref{intro}, most prior works that try to solve 
class-imbalance can
be categorized into 3 domains : (1) Data re-sampling techniques, (2) Metric 
learning and knowledge transfer and (3) Cost-sensitive
learning methods.

\subsection{Data Re-sampling}

Data re-sampling techniques try to balance the number of samples among the
classes by using various sampling techniques during the data pre-processing. The
sampling techniques, used for the purpose, either randomly over-sample data from
the minority classes or randomly under-sample data from the majority classes or
both. Over-sampling from the minority classes \cite{Diversity_based_oversampling,Comparing_oversampling_techniques} replicates the 
available
data samples in order to increase the number of samples. But such a practice 
introduces the risk of overfitting. Synthetic Minority Over-sampling Technique 
(SMOTE)
\cite{SMOTE}, proposed by Chawla et al., increases the number of data samples for the 
minority
classes by creating synthetic data using interpolation among the original data 
points.
Though SMOTE only used the minority class samples while generating data samples,
later variants of SMOTE (e.g., Border-line SMOTE \cite{BorderlineSMOTE} and Safe-level SMOTE \cite{SafelevelSMOTE})
take the majority class samples into consideration as well. But such data 
generation
techniques do not guarantee that the synthesized data points will always follow 
the
actual data distribution of the minority classes. On the other hand, 
under-sampling
techniques \cite{Undersampling_1,Undersampling_2} reduce data from the majority classes and might result in 
cutting
out some important data samples.

\subsection{Metric Learning and Knowledge Transfer}

Metric learning aims to learn an embedding function that can embed data to a 
feature space where the inter-data relationships are preserved. Contrastive 
embedding \cite{contrastive} is learned using paired data samples to minimize 
the distance between the features of same class samples while maximizing the 
distance between different class samples. Song et al. \cite{deepmetriclearning} 
proposed a structured feature embedding based on positive and negative samples 
pairs in the dataset. Triplet loss \cite{tripletloss}, on the other hand, uses 
triplets instead of pairs, where one sample is considered the anchor. Metric 
learning still faces the risk of learning embedding functions biased towards the 
majority classes. Some recent works (e.g., OLTR \cite{OLTR}) have also tried to 
transfer knowledge from the majority classes to the minority classes either by 
meta learning \cite{metalearning1} or by adding an external memory module 
\cite{OLTR}. Even though OLTR \cite{OLTR} performs well for long-tailed 
classification, as pointed out by \cite{decoupling}, their design of external 
memory modules might be a non-trivial and expensive task.

\subsection{Cost-Sensitive Learning}

Cost-sensitive learning techniques try to penalize the DNN higher for making
prediction mistakes on certain data samples than on the others. To achieve 
that, different weights are assigned to different samples and the penalty 
incurred on each
data sample is scaled up/down using the corresponding weight. Research in this 
domain mainly target to find an effective way to assign these weights to the 
samples. To
solve class-imbalance, majority of the works propose techniques that assign 
higher
weights to the minority class samples. Such techniques ensure that the DNN gets
higher penalty for making mistakes on the minority class samples. One such 
simple and commonly used weight distribution technique is to use the inverse 
class-frequencies as the weights for the samples each class 
\cite{Inverse_freq1,Inverse_freq2}. Later 
variants of this
technique \cite{Smoothed_inv} use a smoothed version of the square root of class-frequencies 
for the
weight distribution. Class-balanced loss \cite{Class_balancedloss} proposed by Lin et al. calculates 
the effective number of samples of each class and uses it to assign weights to 
the samples.
All of the above mentioned works assume that the minority classes are always the
most weakly represented classes and therefore needs high weights. But that 
assumption might not always be true because certain minority class might be 
sufficiently
represented by a small number of samples. Giving high weights to the samples of
such classes might cause drop in overall performance. Therefore, Tsung-Yi et al. 
proposed a sample-based weighting technique called “Focal loss” \cite{FOCALloss}, where each 
sample is assigned a weight based on it’s difficulty. The difficulty of each 
sample is quantified in terms of the loss incurred by the DNN on that sample, 
where more lossy samples imply more difficult samples. Though focal loss \cite{FOCALloss}
was originally proposed for
dense object detection tasks, it has also become popular in class-imbalanced 
classification tasks \cite{Class_balancedloss}. The minority classes are expected to have more difficult 
samples compared to the majority classes and therefore get high weights by focal 
loss. Indeed the proportion of difficult samples in a minority class is more 
than that in the majority
class. However, in terms of absolute number of difficult samples, the majority 
class
surpasses the minority class, as it is much more populated than the minority 
class.
Therefore, giving high weights to all difficult samples irrespective of their 
classes still
biases the DNN’s performance.

Our work also lies in the regime of cost-sensitive learning. We propose a 
dynamic weighting system that dynamically assigns weights to each sample of each
class based on the instantaneous difficulty of the class, rather than that of 
each sample, as perceived by the DNN. Our weighting system helps to boost the 
performance
of commonly used loss functions (e.g., cross-entropy loss) in class-imbalanced 
situations.
\section{Proposed Method}

%
%

\subsection{Measuring Class Difficulty}

Human beings use the metric ‘difficulty’ majorly to give a qualitative 
description
of things, for example “this task is very difficult” or “this game is so easy”. 
Similar
behavior can also be seen in neural networks where they find some parts of a 
task
much more difficult to perform compared to the others. For example, while 
training on a multi-class classification task, the classifier will find some classes easier to learn than the others. 
We propose to measure the difficulty of
each class as perceived by the DNN and use it as clue to determine the weights 
for
the samples. But, as difficulty is a qualitative metric, there is no direct way 
to add a
quantitative value to it. Humans tend to classify a task as difficult, if they 
can not perform well in it. We use a similar approach to use the neural 
network’s performance to
measure the difficulty of classes. During training, the neural network’s 
performance
for each class is measured on a validation data set, which is then used to 
calculate the
class-wise difficulty. The neural network’s performance for any class $c$ is 
measured as its classification accuracy on class $c$, $A_c = n_c/N_c$,
where $N_c$ denotes the total number of samples of class $c$ in validation data 
and $n_c$
denotes the number of class $c$ samples in validation data that the model 
classifies
correctly. Then the difficulty of class $c$, $d_c$, is measured as $d_c=1-A_c$.
A neural network’s perception of “how much a class is difficult to learn” 
changes as
the training process of the network progresses. With time, the network’s 
performance for each class improves and as a result, the 
perceived difficulty of
each class also reduces. Therefore, we calculate the class-difficulties as a 
function of
time as well. The difficulty of class $c$ after training time (i.e., time during 
training) $t$
can be calculated as
\begin{equation}
	d_{c,t} = 1 - A_{c,t}  \text{     },
	\label{diff_calc}
\end{equation}
where $A_{c,t}$
is the neural network’s classification accuracy for class $c$ on the validation
data after training time $t$.
\subsection{Difficulty-Based Weight Distribution}

Once the class-wise difficulty is quantified, then it can be used to assign 
weights
to the classes during training. It is fairly obvious that the classes, that are 
difficult to
learn should be given higher weights compared to the easier classes. Therefore 
the
weight for class $c$ after training time $t$ can be calculated as
\begin{equation}
	w_{c,t} = (d_{c,t})^\tau = (1-A_{c,t})^\tau  \text{     },
    \label{weight_calculation}
\end{equation}
where $d_{c,t}$ 
is the difficulty of class $c$ after time $t$ and $\tau$ is a hyper-parameter. 
The
weight distribution $w_t = \{w_{1,t},w_{2,t},\dots,w_{C,t} \}$ over all $C$ classes can be computed by
repeating Equation \ref{weight_calculation} for all classes.

The hyper-parameter $\tau$ is introduced to control how much we down-weight the
samples of the easy classes. Increasing value of $\tau$ relatively increases the 
classifier’s
focus on the difficult classes. 
\begin{figure}[t]
\centering
\includegraphics[scale=0.5]{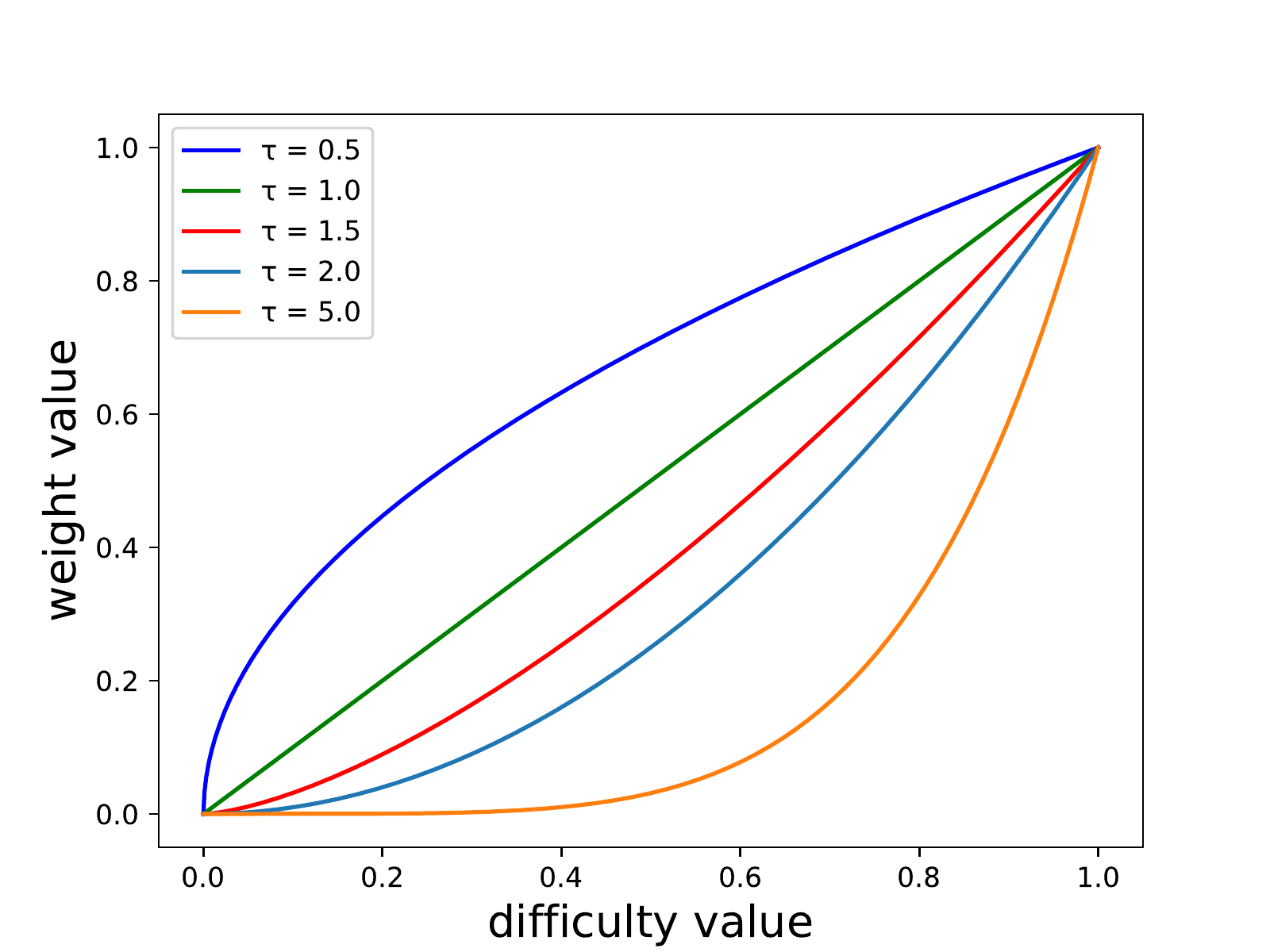}
\caption{
Effect of changing $\tau$ on the difficulty-based weight distribution. Increasing value of
$\tau$ puts heavier weights on the samples of the classes with higher difficulty, while lowering
weights for the easier classes.
}
\label{fig:difficulty_vs_weight}
\end{figure}

Fig. \ref{fig:difficulty_vs_weight} shows how change in the value of $\tau$ 
changes the
weight values for classes of different difficulties. Its effect is almost 
similar to that of the focusing parameter $\gamma$ in focal loss \cite{FOCALloss}. The 
performance of our
proposed method varies significantly with change in value of $\tau$ and the best 
value
for $\tau$ differs from dataset to dataset. Unfortunately, the only way to 
search for the
best value of $\tau$ is by trial and error. To avoid that, we propose a way to 
dynamically
update the value of $\tau$. For dynamically updating $\tau$, the value of $\tau$ 
after training time $t$
is calculated as
\begin{equation}
	\tau_t = \frac{2}{1 + \exp(-b_t)} \text{     },
	\label{tau_calc}
\end{equation}
where $b_t$ measures the bias in the performance of the classifier over $C$ 
classes as
\begin{equation}
	b_t = \frac{\max_{c=1,2,\dots,C} A_{c,t}}{\min_{c'=1,2,\dots,C} A_{c',t} + 
	\epsilon} - 1 \text{     } .
   \label{bias_calc}
\end{equation}
In Equation \ref{bias_calc}, $\epsilon$ is a small positive value (= +0.0001) 
introduced to handle situations
where $\min_{c'=1,2,\dots,C} A_{c',t} = 0 $. Equation \ref{tau_calc} increases 
the value of $\tau$ when the classification performance of the classifier is 
highly biased (i.e., high $b_t$) and decreases it in
case of low bias (i.e., less $b_t$).

\subsection{Class-Wise Difficulty-Balanced Softmax Cross-Entropy Loss}
Suppose when an input data is fed to the classifier after training time $t$ 
during
training, the predicted output of the classifier for all $C$ classes are $z_t = 
\{z_{1,t}, z_{2,t},\dots, z_{C,t} \}$. 
The probability distribution 
$p_t=\{p_{1,t},p_{2,t},...,p_{C,t} \}$ over all the classes is computed using 
the
softmax function, which is
\begin{equation}
	p_{j,t} = \frac{\exp{z_{j,t}}}{\sum_{i=1}^{C} \exp{z_{i,t}}}  \text{     }
	\forall j \in 1,2,\dots,C \text{ }.
\end{equation}

For an input data sample of class $k$, cross-entropy (CE) loss function computes 
the
loss after training time $t$ as
\begin{equation}
	\text{CE}(p_t, k) = -\log p_{k,t} \text{     } .
\end{equation}
For the same input data sample, our class-wise difficulty-balanced softmax 
cross-entropy (CDB-CE) loss function computes the loss after training time $t$ 
as
\begin{equation}
	\text{CDB-CE}(w_t,p_t,k) = -w_{k,t} \log p_{k,t} \text{     }.
\end{equation}
To make the weights time-dependent, we calculate them 
after
each epoch using the model’s class-wise validation accuracy.
\section{Experiments}
To demonstrate our proposed solution’s ability to generalize to any data-type or
dataset, we evaluate the effectiveness of our solution on 4 different datasets 
namely
MNIST, long-tailed CIFAR, ImageNet-LT and EGTEA. MNIST, long-tailed CIFAR and ImageNet-LT are image 
datasets while EGTEA is a video dataset.

\subsection{Datasets}
\textbf{MNIST.} From the standard MNIST handwritten digit recognition 
dataset \cite{MNIST_data}, we
generate a class-imbalanced binary classification task using a subset of the 
dataset.
The experimental setup is exactly same as given in \cite{reweighting_examples}. We select a total of 
5000 training images of class ‘4’ and ‘9’ where ‘9’ is chosen as the majority 
class. We calculate
the ‘majority class ratio’ as
\begin{equation}
	\text{majority class ratio} = \frac{\text{no. of training samples in majority 
	class}}{\text{no. of training samples}} \text{     }.
    \label{majority_class_ratio}
\end{equation}
Increasing the majority class ratio increases the imbalance in the training 
dataset. We also use a validation set which is created by selecting 500 images 
for each of the two classes from the original dataset but these images are 
different from the 5000 images selected for training. A test set was 
also created by randomly selecting 800 images for each of the classes from the 
original MNIST test set.

\textbf{Long-Tailed CIFAR.} We conduct experiments on long-tailed 
CIFAR-100 \cite{CIFAR}. First a validation set was created from the original training set 
by randomly selecting 50 images per class. After the separation of the 
validation set, the remaining images in the training set were used to create a 
long-tailed version of the dataset using the exact same procedure as stated in 
\cite{Class_balancedloss}. The number of training images per class are reduced following an 
exponential function $n=n_c \mu^c$, where $c$ is the 0-based index of the class 
and $n_c$ is the remaining number of training images of class $c$ after 
separation of the validation set and $\mu \in (0,1)$.  Similar to \cite{Class_balancedloss}, the 
‘imbalance’ factor of a dataset is defined as the number of training samples in 
the largest class divided by that of the smallest class. The test set used for 
experiment is exactly same as the original CIFAR test set available and is a 
balanced set.

\textbf{ImageNet-LT.} We also conduct experiments on the long-tailed version of the original ImageNet-2012 \cite{Imagenet}, as constructed in \cite{OLTR}. It comprises of 115,800 images from 1000 categories, where the most frequent class has 1280 image samples and the least frequent class has only 5 images. The test set is balanced. The split constructed by \cite{OLTR} also provides a validation set that is separate from the test set and training set.

\textbf{EGTEA.} We also conduct experiments on the EGTEA Gaze+ dataset 
\cite{EGTEA}. This is an egocentric dataset that contains trimmed video clips of many 
kitchen-related actions. These video clips are extracted segments from longer 
videos that were collected by 32 different subjects. Each video clip is 
assigned 
a single action label and the challenge is to train a classifier to classify the 
actions from the provided video clips. Each action label is made up of a verb 
and a set of nouns (e.g., the action ‘Wash Plate’ is made up of the verb ‘Wash’ 
and noun ‘Plate’). The noun classification task is very similar to the image 
classification task. So, for our experiments, we focus on the
verb classification in order to test our proposed method on diverse tasks. This 
dataset has 19 different verb classes (e.g., ‘Open’, ‘Wash’ etc.). EGTEA Gaze+ 
\cite{EGTEA} is inherently class-imbalanced. Fig. \ref{fig:egteafreq}(a) shows 
the data distribution of the 
EGTEA dataset. For our experiments, we use the split1 of the EGTEA dataset (8299 
training video clips and 2022 testing video clips). We create our validation set 
by using the training video clips from subjects P20 to P26, resulting in 1927 
validation video clips and 6372 training clips.

\subsection{Implementation Details}
We use a LeNet-5 \cite{LeNet} model for MNIST unbalanced binary classification 
experiments following \cite{reweighting_examples}. The model is trained for $4000$ epochs on a single 
NVIDIA GeForce GTX 1080 GPU with a batch size of $100$. As optimizer, we use 
Stochastic Gradient Descent (SGD) with momentum of $0.9$, weight decay of 
$0.0005$ and an initial learning rate of $0.001$, which is decayed by $0.1$ 
after $3000$ 
epochs. The trained model is tested on the balanced test set.

For experiments on long-tailed CIFAR, we follow the exact same implementation
strategy as provided in \cite{Class_balancedloss}. We train a ResNet-32 \cite{ResNet} model for $200$ epochs 
using a batch size of $128$ on $4$ NVIDIA Titan X GPUs. We use SGD optimizer with 
momentum $0.9$ and weight decay of $0.0005$. An initial learning rate of $0.1$ 
is 
used, which is decayed by $0.01$ after $160$ and $180$ epochs.

For experiments on ImageNet-LT, we use the same setup as in \cite{OLTR}. We use ResNet-10 \cite{ResNet} model for the purpose. The trained model is tested on the balanced test data. 

For EGTEA dataset, we use a 3D-ResNeXt101 \cite{ResNeXt,3DResNext} model and train it for $100$
epochs on $8$ NVIDIA Titan X GPUs using a batch size of $32$. SGD with momentum 
$0.9$
is used with a weight decay of $0.0005$ and an initial learning rate of $0.001$, 
which is decayed by $0.1$ after $60$ epochs. During training, we sample $10$ RGB 
frames from each video clip by dividing the clip into $10$ equal segments 
followed 
by randomly selecting one RGB frame from each segment. We use random-cropping, 
random-rotating and horizontal-flipping as data augmentation. Training input 
size is $10 \times 3 \times 224 \times 224$.
During testing and validation, we sample $10$ RGB frames at equal intervals from 
each video clip. 

We use PyTorch \cite{Pytorch} framework for all our implementations. For all datasets, our CDB-CE loss implementation calculates the class-wise weights after every epoch using the model's 
class-wise validation accuracy. 

\subsection{Results on Unbalanced MNIST Binary Classification}
\label{MNIST_results}
Similar to \cite{reweighting_examples}, we increase the majority class ratio defined in Equation 
\ref{majority_class_ratio} from
$0.9$ to $0.995$ by increasing the number of training samples of majority class, 
while
keeping the total number of training samples constant at $5000$. Following 
implementation details of \cite{reweighting_examples}, we retrain LeNet-5 \cite{LeNet} for each majority class 
ratio using different loss functions and compare the error rate of the trained 
model on the
test set. Table \ref{table:MNIST1} compares the effect of increasing majority class ratio on the 
test error rates of LeNet-5, trained using various weighted and unweighted loss 
functions.
For comparison, we use the mean and standard deviation of the classification 
error
rates achieved over 10 runs using random splits. The compared loss functions 
include (1) Unweighted Cross-Entropy(CE) , which uses an unweighted softmax 
cross-entropy loss function to train the model; (2) inverse class-frequency 
weighting (IFW) \cite{Inverse_freq2}, which uses a weighted softmax cross-entropy loss function where the weight 
for
each class is calculated using the inverse of it’s frequency; (3) Focal loss \cite{FOCALloss}, ClassBalanced(CB) loss \cite{Class_balancedloss}, Equalization loss (EQL) \cite{Equalizationloss} and L2RW \cite{reweighting_examples} are
state-of-the-art loss functions.
\setlength{\tabcolsep}{4pt}
\begin{table}[t!]
\begin{center}
\caption{
Mean and standard deviation of classification error rates (\%) of LeNet-5 
\cite{LeNet} trained
for MNIST \cite{MNIST_data} imbalanced binary classification using different 
loss functions for different majority class ratios. Here we show the best 
results obtained by each of the loss functions in our
implementation. For class-wise difficulty-balanced softmax cross-entropy (CDB-CE) loss (Ours), we report the results with dynamically updated $\tau$.
}
\label{table:MNIST1}
\begin{tabular}{llllll}
\hline\noalign{\smallskip}
Maj. class ratio & 0.9 & 0.95 & 0.98 & 0.99 & 0.995\\
\noalign{\smallskip}
\hline
\noalign{\smallskip}
Unweighted CE &$1.50 \pm 0.51$ & $2.36 \pm 0.46$ & $5.09 \pm 0.41$ & $8.59\pm 0.41$ & $14.35 \pm 1.10$\\
IFW \cite{Inverse_freq2}  &$1.16 \pm 0.40$ & $1.74 \pm 0.31$ & $3.13 \pm 0.74$ & $6.01\pm 0.56$ & $8.94 \pm 0.70$\\
Focal Loss \cite{FOCALloss}  &$1.74 \pm 0.26$ & $2.78 \pm 0.29$ & $6.67 \pm 0.63$ & $11.11\pm 1.20$ & $17.17 \pm 0.86$\\
CB Loss \cite{Class_balancedloss}&$1.07 \pm 0.23$ & $1.79 \pm 0.39$ & $3.58 \pm 0.71$ & $5.88\pm 1.20$ & $8.61 \pm 1.11$\\
EQL \cite{Equalizationloss} &$1.49 \pm 0.34$ & $2.26 \pm 0.41$ & $2.43 \pm 0.14$ & $2.60\pm 0.33$ & $3.71 \pm 0.41$\\
L2RW \cite{reweighting_examples} &$1.24 \pm 0.69$ & $1.76 \pm 1.12$ & $2.06 \pm 0.85$ & $2.63\pm 0.65$ & $3.94 \pm 1.23$\\
CDB-CE(Ours) &$\mathbf{0.74 \pm 0.14}$ & $\mathbf{1.27\pm 0.33}$ & $\mathbf{1.65 \pm 0.26}$ & $\mathbf{2.39\pm 0.41}$ & $\mathbf{3.71 \pm 0.27}$\\
\hline
\end{tabular}
\end{center}
\end{table}
\setlength{\tabcolsep}{1.4pt}

As can be seen from Table \ref{table:MNIST1}, our class-wise difficulty-balanced cross-entropy
(CDB-CE) loss function performs better than the others. But to ensure a good 
performance, it is important to select an appropriate value of $\tau$. Hence we 
conduct another experiment to investigate the effect of changing $\tau$ on the 
performance of our
method. For that, we compare the performance of our CDB-CE loss function for 
different values of $\tau$ and different majority class ratios. The results of 
the experiment
are listed in Table \ref{table:MNIST2}.

\setlength{\tabcolsep}{4pt}
\begin{table}[t!]
\begin{center}
\caption{
Mean and standard deviation of classification error rates (\%) of LeNet-5 
\cite{LeNet} trained
for MNIST imbalanced binary classification using our CDB-CE loss with different values of $\tau$
For dynamically updating $\tau$ (last row), we update the value of $\tau$ after every epoch as given in Equation \ref{tau_calc}.}
\label{table:MNIST2}
\begin{tabular}{llllll}
\hline\noalign{\smallskip}
Maj. class ratio & 0.9 & 0.95 & 0.98 & 0.99 & 0.995\\
\noalign{\smallskip}
\hline
\noalign{\smallskip}
$\tau = 0.5$ &$1.06 \pm 0.34$ & $1.43\pm 0.24$ & $1.93 \pm 0.27$ & $2.59\pm 0.44$ & $4.03 \pm 0.41$\\
$\tau = 1.0$ &$0.90 \pm 0.27$ & $1.38\pm 0.20$ & $1.86 \pm 0.30$ & $2.49\pm 0.57$ & $3.94 \pm 0.36$\\
$\tau = 1.5$ &$0.85 \pm 0.19$ & $1.35\pm 0.31$ & $1.71 \pm 0.28$ & $2.31\pm 0.38$ & $\mathbf{3.54 \pm 0.25}$\\
$\tau = 2.0$ &$0.75 \pm 0.15$ & $1.21\pm 0.32$ & $1.75 \pm 0.36$ & $\mathbf{2.23\pm 0.34}$ & $3.65 \pm 0.41$\\
$\tau = 5.0$ &$0.88 \pm 0.25$ & $\mathbf{1.19\pm 0.36}$ & $2.00 \pm 0.32$ & $2.51\pm 0.41$ & $3.78 \pm 0.43$\\
$\tau = 7.0$ &$0.96 \pm 0.20$ & $1.20\pm 0.20$ & $2.04 \pm 0.30$ & $2.64\pm 0.40$ & $4.13 \pm 0.37$\\
dyn. updated $\tau$ &$\mathbf{0.74\pm 0.14}$ & $1.27\pm 0.33$ & $\mathbf{1.65 \pm 0.26}$ & $2.39\pm 0.41$ & $3.71 \pm 0.27$\\
\hline
\end{tabular}
\end{center}
\end{table}
\setlength{\tabcolsep}{1.4pt}

As can be seen from Table \ref{table:MNIST2}, increasing value of $\tau$ initially helps in 
improving the performance of our method but after a certain point, it leads to a 
drop in the performance. We believe that the drop comes due to the excessive 
down-weighting of the samples of the easy classes. In Table \ref{table:MNIST2}, our method works 
well over a wide range of $\tau$ values. The best value for $\tau$ varies even 
with the majority class ratio. But one interesting thing to notice is that even 
though dynamically updating $\tau$ does not always give
the best performance, it’s performance is never too far from the best and it 
consistently outperforms all the existing methods listed in Table 1. Therefore, 
dynamically updating $\tau$ can be a default choice to select the value of $\tau
$ in case one wants to avoid trial and error searching for the best $\tau$.

\subsection{Results on Long-Tailed CIFAR-100}
We conduct extensive experiments on long-tailed CIFAR-100 dataset 
\cite{CIFAR,Class_balancedloss} as
well. ResNet-32 \cite{ResNet} is retrained for different imbalance factors in the 
training dataset, using different loss functions. Table \ref{table:CIFAR1} reports the 
classification accuracy(\%) of each such trained model on the CIFAR-100 test 
set. We compare the results of our method with that of Focal loss \cite{FOCALloss}, Class-Balanced loss \cite{Class_balancedloss}, L2RW \cite{reweighting_examples}, Meta-Weight Net \cite{Meta_net_weight} and Equalization loss \cite{Equalizationloss}.

\setlength{\tabcolsep}{4pt}
\begin{table}[tp]
\begin{center}
\caption{
Top-1 classification accuracy (\%) of ResNet-32 trained on long-tailed CIFAR-100
training data. $\dagger$ means that the result has been copied from the origin 
paper \cite{Class_balancedloss,Meta_net_weight,LDAM-DRW}. For CDB-CE loss 
(Ours), we report the results with dynamically updated $\tau$.
}
\label{table:CIFAR1}
\begin{tabular}{llllll}
\hline\noalign{\smallskip}
Imbalance & 200 & 100 & 50 & 20 & 10\\
\noalign{\smallskip}
\hline
\noalign{\smallskip}
Focal loss $\dagger$ \cite{FOCALloss} &$35.62$ & $38.41$ & $44.32$ & $51.95$ & $55.78$\\
Class-Balanced $\dagger$ \cite{Class_balancedloss} &$36.23$ & $39.60$ & $45.32$ & $52.99$ & $57.99$\\
L2RW $\dagger$ \cite{reweighting_examples}&$33.38$ & $40.23$ & $44.44$ & $51.64$ & $53.73$\\
Meta-Weight Net $\dagger$ \cite{Meta_net_weight}&$\mathbf{37.91}$ & $42.09$ & $46.74$ & $\mathbf{54.37}$ & $58.46$\\
Equalization loss\footnotemark \footnotesize $\dagger$ \cite{Equalizationloss} &$37.34$ & $40.54$ & $44.70$ & $54.12$ & $58.32$\\
LDAM-DRW $\dagger$ \cite{LDAM-DRW} &  -- & 42.04 &   -- &  -- & 58.71\\
CDB-CE (Ours) &$37.40$ & $\mathbf{42.57}$ & $\mathbf{46.78}$ & $54.22$ & $\mathbf{58.74}$\\
\hline
\end{tabular}
\end{center}
\end{table}
\setlength{\tabcolsep}{1.4pt}

As can be seen from Table \ref{table:CIFAR1}, our CDB-CE loss with dynamically 
updated $\tau$ provides better performance than
the others in most cases. But as stated in Section \ref{MNIST_results}, the 
results of our method depend 
highly on the value of $\tau$. Hence, we conduct a further study of how the 
performance of our method varies on the long-tailed CIFAR-100 dataset with the 
change in $\tau$. The results are shown in Table \ref{table:CIFAR2}.

\setlength{\tabcolsep}{4pt}
\begin{table}[t]
\begin{center}
\caption{
Top-1 classification accuracy (\%) of ResNet-32 trained using class-wise difficulty-balanced cross-entropy (ours) loss function for different values of $\tau$. $\tau = 0$ means the original
unweighted softmax cross-entropy loss function.
}
\label{table:CIFAR2}
\begin{tabular}{llllll}
\hline\noalign{\smallskip}
Imbalance & 200 & 100 & 50 & 20 & 10\\
\noalign{\smallskip}
\hline
\noalign{\smallskip}
$\tau=0$ &$34.95$ & $38.21$ & $43.89$ & $51.34$ & $55.65$\\
$\tau=0.5$ &$37.21$ & $41.26$ & $46.13$ & $\mathbf{54.60}$ & $58.29$\\
$\tau=1.0$ &$\mathbf{37.99}$ & $41.67$ & $46.45$ & $53.48$ & $\mathbf{59.47}$\\
$\tau=1.5$ &$37.63$ & $\mathbf{42.70}$ & $\mathbf{47.09}$ & $52.74$ & $58.68$\\
$\tau=2.0$ &$37.03$ & $42.44$ & $46.94$ & $53.00$ & $58.65$\\
$\tau=5.0$ &$36.81$ & $40.62$ & $45.45$ & $51.67$ & $54.48$\\
dynamically updated $\tau$ &$37.40$ & $42.57$ & $46.78$ & $54.22$ & $58.74$\\
\hline
\end{tabular}
\end{center}
\end{table}
\setlength{\tabcolsep}{1.4pt}

Using $\tau = 0$ makes our CDB-CE loss drop back to the original unweighted 
softmax cross-entropy loss function. From Table \ref{table:CIFAR2}, almost a wide range of values 
for $\tau$
helps our weighted loss to get better results than the baseline of $\tau = 0$. 
Again the interesting thing is even though dynamically updating $\tau$ does not 
give the best results, it’s performance is not far from the best and it 
outperforms existing methods of Table \ref{table:CIFAR1} in majority of the cases.
\footnotetext{The equalization loss results reported in \cite{Equalizationloss} use more augmentation techniques (e.g.,
Cutout \cite{Cutout}, autoAugment \cite{Autoaugment}) compared to \cite{Class_balancedloss,Meta_net_weight}. Hence for fair comparison, we report
the results that we achieved without using the additional augmentation.}

\subsection{Results on ImageNet-LT}
We compare the performance of our method on ImageNet-LT with other state-of-the-art methods. For comparison, we use the top-1 classification accuracy as our evaluation metric. The results are listed in the Table \ref{table:ImageNet1}.
\setlength{\tabcolsep}{4pt}
\begin{table}[tp]
\begin{center}
\caption{
Top-1 classification accuracy (\%) of ResNet-10 on ImageNet-LT for different methods. $\dagger$ means that the result has been copied from the origin paper \cite{Equalizationloss,OLTR,decoupling}.
}
\label{table:ImageNet1}
\begin{tabular}{ll}
\hline\noalign{\smallskip}
Method & Top-1 Accuracy(\%)\\
\noalign{\smallskip}
\hline
\noalign{\smallskip}
Focal loss $\dagger$ \cite{FOCALloss} &$30.50$\\
OLTR $\dagger$ \cite{OLTR} &$35.60$\\
Joint training $\dagger$ \cite{decoupling} & 34.80\\
Equalization loss $\dagger$ \cite{Equalizationloss} &$36.44$\\
OLTR \cite{OLTR} + CDB-CE(Ours) & $36.70$\\
Joint training \cite{decoupling}+ CDB-CE(Ours)& $37.10$\\
CDB-CE (Ours) &$\mathbf{38.49}$\\
\hline
\end{tabular}
\end{center}
\end{table}
\setlength{\tabcolsep}{1.4pt}

For OLTR \cite{OLTR}+CDB-CE and Joint training \cite{decoupling}+CDB-CE, we 
implemented our method in the original implementations of \cite{OLTR,decoupling} 
available on github. From Table \ref{table:ImageNet1}, it can be seen that our 
CDB-CE loss not only achieves the best result but it also helps to boost the 
performance of OLTR and Joint training. 

\subsection{Results on EGTEA}
We also conduct extensive experiments on EGTEA \cite{EGTEA} dataset. As shown in 
Fig. \ref{fig:egteafreq}, EGTEA dataset is inherently class-imbalanced. The average amount of training
samples per class is 1216.0 for the five most frequent classes while for the 
rest of the 14 classes, the average is only 158.5. That is why we define the 
five most frequent classes (i.e., ‘Take’ , ‘Put’ , ‘Open’ , ‘Cut’ and ‘Read’ ) 
together as our ‘majority classes’ and the rest of the classes as our ‘minority 
classes’. Table \ref{table:EGTEA1} reports the results on the
test split of a 3D-ResNeXt101, trained on EGTEA dataset using various loss 
functions. For comparison, we use four different metrics (1) ‘Acc@Top1’ is the 
micro-average of the top-1 accuracies of all the classes (2) ‘Acc@Top5’ is the 
micro-average of the top-5 accuracies of all the classes (3) ‘Recall’ is the 
macro-average of the recall values of all the classes (4) ‘Precision’ is the 
macro-average of the precision values of all the classes.

\setlength{\tabcolsep}{4pt}
\begin{table}[t]
\begin{center}
\caption{
3D-ResNeXt101 results on EGTEA test set. For class-wise difficulty-balanced cross
entropy (ours), we use dynamically updated $\tau$.
}
\label{table:EGTEA1}
\begin{tabular}{lllll}
\hline\noalign{\smallskip}
 & Acc@Top1 & Acc@Top5  & Recall & Precision \\
\noalign{\smallskip}
\hline
\noalign{\smallskip}
Unweighted CE &$67.41$ & $95.40$ & $64.77$ & $61.73$\\
Focal loss \cite{FOCALloss} &$64.34$ & $94.36$ & $59.17$ & $59.09$\\
Class-Balanced loss \cite{Class_balancedloss}&$66.86$ & $95.69$ & $63.26$ & $63.39$\\
CDB-CE (Ours)&$\mathbf{69.14}$ & $\mathbf{96.84}$ & $\mathbf{66.24}$ & $\mathbf{63.86}$\\
\hline
\end{tabular}
\end{center}
\end{table}
\setlength{\tabcolsep}{1.4pt}

From Table \ref{table:EGTEA1}, our proposed method achieves significant performance gains in
all the metrics compared to other loss functions. In order to ensure that these 
gains are not entirely because of the majority classes, we conduct a further 
study, where we compare the performance of different loss functions on the 
‘majority classes’ and ‘minority classes’ separately. The results are tabulated 
in Table \ref{table:EGTEA2}.

\setlength{\tabcolsep}{4pt}
\begin{table}[t]
\begin{center}
\caption{
3D-ResNeXt101 results on the ‘majority classes’ and ‘minority classes’ for different
training loss functions. As explained before, the five most frequent classes together constitute
the ‘majority classes’ while the rest of them are the ‘minority classes’. We use ‘Recall’ and ‘Precision’ for comparison.
}
\label{table:EGTEA2}
\begin{tabular}{lllll}
\hline\noalign{\smallskip}
 & \multicolumn{2}{l}{Majority classes} &  \multicolumn{2}{l}{Minority classes} \\
\noalign{\smallskip}
\hline
\noalign{\smallskip}
 & Recall & Precision & Recall & Precision \\
\noalign{\smallskip}
\noalign{\smallskip}
Unweighted CE &$74.91$ & $\mathbf{75.62}$ & $61.14$ & $56.75$\\
Focal loss \cite{FOCALloss} &$70.27$ & $75.00$ & $55.21$ & $53.40$\\
Class-Balanced loss \cite{Class_balancedloss}&$\mathbf{75.95}$ & $73.75$ & $58.72$ & $59.68$\\
CDB-CE (Ours)&$74.42$ & $73.51$ & $\mathbf{63.31}$ & $\mathbf{60.42}$\\
\hline
\end{tabular}
\end{center}
\end{table}
\setlength{\tabcolsep}{1.4pt}

Table \ref{table:EGTEA2} confirms that our proposed method helps to improve the macro-averaged
recall and precision on the ‘minority classes’. Though we see a drop in average 
recall and precision for the ‘majority classes’ using our method, Table \ref{table:EGTEA1} shows 
that we achieve an overall performance gain for both precision and recall. 
Therefore, improvement on the ‘minority classes’ accounts for the overall gain.


\section{Conclusion}
In this paper, we have proposed a new weighted-loss method for solving 
class-imbalance. The key idea of our method is to take the difficulty of each 
class into consideration, rather than the number of training samples of the 
class, for assigning weights to samples. Based on this idea, we define a 
quantification for the dynamic difficulty of each class. Further we propose a 
difficulty-based weighting system that dynamically assigns weights to the 
samples based on the difficulty of their classes. We also conduct extensive 
experiments on artificially induced class-imbalanced MNIST, CIFAR and ImageNet datasets 
and inherently class-imbalanced EGTEA dataset. The experimental results show 
that using our weighting strategy with cross-entropy loss function helps to 
boost its performance and achieve best results on imbalanced
datasets. Moreover, achieving good results on both image and video datasets show
that the benefit of our method is not limited to any particular type of data.
\bibliographystyle{splncs}
\bibliography{egbib1}

\end{document}